\documentclass[sigconf]{acmart}
\usepackage{bm}
\usepackage{amsmath,graphicx,bbm,color,booktabs}
\usepackage{amsopn}

\DeclareMathOperator*{\argmax}{argmax}

\renewcommand{\eqref}[1]{Eq.~(\ref{#1})}

\usepackage{multirow}
\AtBeginDocument{%
  \providecommand\BibTeX{{%
    \normalfont B\kern-0.5em{\scshape i\kern-0.25em b}\kern-0.8em\TeX}}}

\copyrightyear{2022}
\acmYear{2022}
\setcopyright{acmcopyright}\acmConference[MM '22]{Proceedings of the 30th ACM International Conference on Multimedia}{October 10--14, 2022}{Lisboa, Portugal}
\acmBooktitle{Proceedings of the 30th ACM International Conference on Multimedia (MM '22), October 10--14, 2022, Lisboa, Portugal}
\acmPrice{15.00}
\acmDOI{10.1145/3503161.3548340}
\acmISBN{978-1-4503-9203-7/22/10}

\begin{document}

\title{A Baseline for Detecting Out-of-Distribution Examples in Image Captioning}

\author{Gabi Shalev}
\authornotemark[1]
\affiliation{%
  \institution{Bar-Ilan University}
  \country{Israel}
}
\email{shalev.gabi@gmail.com}

\author{Gal-Lev Shalev}
\authornote{Both authors contributed equally to this research.}

\affiliation{%
  \institution{Bar-Ilan University}
  \country{Israel}
}
\email{gallev898@gmail.com}

\author{Joseph Keshet}
\affiliation{%
  \institution{Technion}\country{Israel}
}
\email{jkeshet@technion.ac.il}


\begin{abstract}
Image captioning research achieved breakthroughs in recent years by developing neural models that can generate diverse and high-quality descriptions for images drawn from the same distribution as training images. However, when facing out-of-distribution (OOD) images, such as corrupted images, or images containing unknown objects, the models fail in generating relevant captions.

  In this paper, we consider the problem of OOD detection in image captioning. We formulate the problem and suggest an evaluation setup for assessing the model's performance on the task. Then, we analyze and show the effectiveness of the caption's likelihood score at detecting and rejecting OOD images, which implies that the relatedness between the input image and the generated caption is encapsulated within the score. 
\end{abstract}

\ccsdesc[500]{Computing methodologies~Computer vision tasks}
\keywords{out-of-distribution detection, image captioning, uncertainty estimation, anomaly detection}

\maketitle

\section{Introduction}

\begin{figure}[th!]
  \centering
  \centerline{\includegraphics[width=7cm]{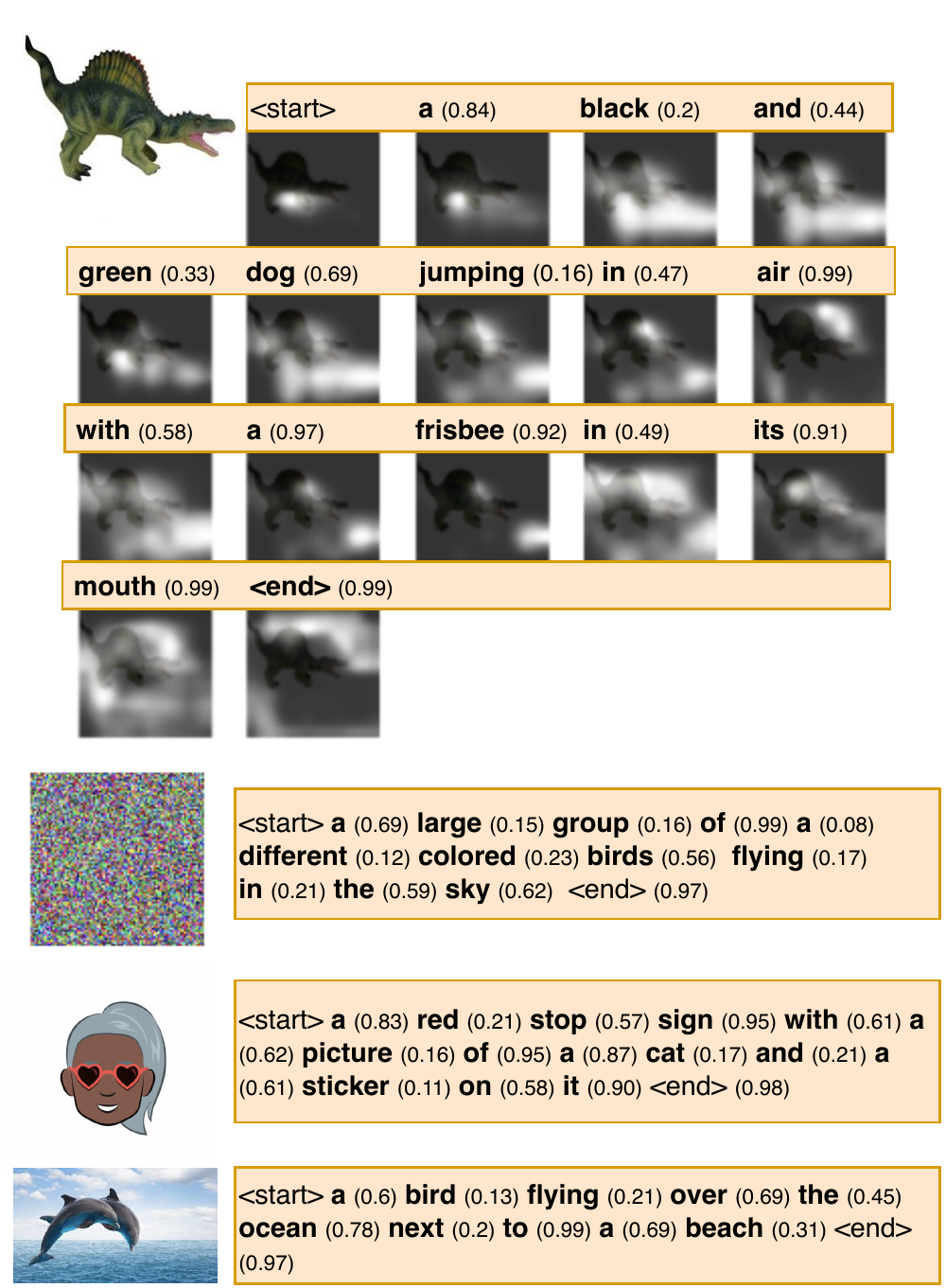}}
  \caption{Generated captions for out-of-distribution images by a Top-Down model trained on COCO dataset. The first image contains an object that does not appear in training (unknown object), without a scene in the background. Second is an image containing a random noise. Third is an image from \textit{Google's cartoon set}, and forth is an image of unknown object (\textit{dolphin}) in a natural scene. Attention maps are added for the first example, and the predicted probability of each token is mentioned in brackets.}
  \label{fig:ood_samples}
\end{figure}

Deep Neural Networks (DNNs) have gained lots of success after enabling several breakthroughs in notably challenging problems such as image classification, object detection, and language generation. 
However, despite the tremendous progress driven by DNNs, they were found as vulnerable to OOD inputs. For example, image classification models were shown to fail by predicting the wrong class with high confidence when given images containing unknown objects \cite{nguyen2015deep}. Another example would be in the image captioning (IC) task, where the models are able to output high-quality and diverse sentences when given images drawn from the same distribution as the training set; however, when facing images in the wild, these models are shown to generalize poorly \cite{tran2016rich}.

OOD detection task has been formulated as the task of detecting whether input data is drawn from a distribution different from the training distribution. The task has been studied for a long, and recently, a baseline for detecting OOD images by neural image classification models has been proposed in \cite{hendrycks2016baseline}. The authors investigated the OOD detection rate of image classification models by considering the confidence of the predicted class and set baseline results for the task. The baseline became widely adopted, and many advanced approaches have been suggested for the task \cite{koner2021oodformer,hsu2020generalized, lee2018simple,li2021k, shalev2018out}. Yet, despite the field's progress, detecting OOD images remains a difficult problem, as they are not limited to some particular type of images. Instead, OOD images can have infinite forms, such as images of objects that do not appear in the training set (unknown objects), corrupted images of known objects, or even random noise images.

We focus on detecting OOD instances in image captioning (IC). While in the image classification task, a model is given an image and is expected to correctly classify it to one of $K$ possible classes, in IC, the model is expected to generate a sentence that describes the scene in the image. Since most OOD image detection research works considered the detection in an image classification task environment, adopting these methods to IC is not straightforward. A possible way to reject OOD images in the IC task environment is by designing a two-phase system. First, a pre-trained image classifier is applied to reject OOD images, and then, if the image was not rejected, it is fed into the IC model to generate a caption. However, such a method has several caveats: (1) It requires training and maintaining two separate models, which adds to the time and space complexity. (2) When the training data is composed of only images and their corresponding captions, training an image classifier can be difficult as it requires heuristics for generating a ground-truth class. 
But while these caveats are manageable, there is a bigger question of whether image-classification-based models are applicable to images used to train image captioning models, as the images are different from those used to train image classification models. For example, while image classification models are given images containing a single main dominant object, IC models are given images containing complex scenes with multiple objects interacting with each other.

This paper proposes a simple yet effective method for detecting OOD instances by IC systems. Using an empirical study on several widespread IC models, we show that the generated caption's likelihood score can capture the relatedness between the image and the generated caption and effectively allow the rejection of OOD images. Furthermore, we find that while image-classification-based methods express their uncertainty using a single score (such as the predicted class probability), IC models express their uncertainty majorly in visually grounded tokens (such as Nouns, Verbs, and Adjectives) and outperform in detecting OOD images with a complex scene. Moreover, to facilitate OOD detection research in IC, we construct benchmarks on top of widely used image datasets and set baseline results for this task.

As the research in IC achieved breakthroughs in the field by developing end-to-end IC models that are able to output high-quality and diverse sentences, an increasing number of industrial companies integrated IC models into their applications \cite{dognin2020image, sadeh2019generating}. However, in real-world scenarios, it is most likely that input instances will contain objects that the model did not see during training, which will result in unexpected behavior by the model. A research work \cite{macleod2017understanding}, explored how blind and visually impaired people experience generated image captions. The authors showed that people trust
incorrect AI-generated captions, filling in details to reconcile incongruencies rather than suspecting the caption may be wrong. Therefore, with the increasing adoption of IC models in real-life applications, it is crucial to focus on rejecting OOD inputs without outputting the generated captions for these instances.

Neural generative IC models are encoder-decoder networks, trained end-to-end, and achieve state-of-the-art results in the field. Training the models is commonly done by maximum likelihood estimation (MLE), which maximizes the probability of a ground-truth sentence given its corresponding training image. During inference time, a decoding strategy is applied to search and output the most probable describing sentence for the input image. These training and inference procedures allow models to have rich image descriptiveness ability and achieve high scores in IC metrics (such as Cider \cite{vedantam2015cider} and BLEU \cite{papineni2002bleu}) when given in-distribution images, but their ability of \textit{"expressing"} their uncertainty for OOD images is yet unknown. 

In Fig-\ref{fig:ood_samples} we show captions generated by Top-Down model \cite{anderson2018bottom} for several different types of OOD images. The figure shows that the sentences are vibrant and natural but also loosely related to the given image. Ideally, when the IC model is facing an OOD image that it cannot describe, the decoded sentence should be assigned with a low probability to allow the rejection of the instance. However, MLE training does not guarantee to capture the relatedness between the generated caption and the given image. Here, we explore the effectiveness of the captions' likelihood at expressing uncertainty in the caption when given OOD images. 

In summary, the contribution of this paper is as follows:
\begin{itemize}
    \item We define the task of out-of-distribution detection in image captioning and explore the performance of several top-performing image captioning models.
    \item We construct benchmarks that include multiple types of out-of-distribution image sets and suggest evaluation metrics.
    \item We demonstrate the efficiency of the generated captions' likelihood at detecting out-of-distribution images without requiring any additional data or external knowledge and show that the uncertainty is expressed through the probability of the visually grounded tokens.
\end{itemize}
\section{Related Work}\label{lbl:related_work}
\paragraph{Out-of-distribution detection} Since
DNNs are ubiquitous, present in nearly all segments of the technology industry, from search engines \cite{sadeh2019joint} to critical applications such as self-driving cars \cite{do2018real} and healthcare \cite{shahid2019applications, granovsky2018actigraphy}, it becomes critical to design models that can express uncertainty when
predicting OOD inputs.
OOD detection task has been studied for a long, and recently, a baseline for detecting OOD images by neural image classification models has been proposed in \cite{hendrycks2016baseline}. The authors examined the OOD detection performance of image classification models using a confidence score derived from the network (max softmax probability). Later, several studies proposed improving the baseline by modifying the model's architecture \cite{shalev2018out, koner2021oodformer}, inference procedure \cite{liang2017enhancing, hsu2020generalized} and training criterion \cite{lee2017training, lee2018simple}.

OOD detection methods were also developed for safety-critical image applications \cite{venkatakrishnan2020self}, and also into other modalities such as natural language processing (NLP) and speech processing.
\cite{hendrycks2020pretrained} explored the task of OOD text detection in NLP classification tasks (sentiment analysis and textual entailment). The authors showed that the output confidence scores of pre-trained Transformers can effectively indicate whether the sample is OOD. They also concluded that pre-trained Transformers have an improved OOD detection rate compared to RNN and convolution-based text classification models. More recently, \cite{li2021k} suggested improving the OOD detection by combining a self-supervised training method with an ensemble of text classifiers, and the method sets the state-of-the-art results.

\paragraph{Novel object description} Some of the first attempts at IC \cite{farhadi2010every,kuznetsova2012collective} relied heavily on the outputs of object detectors and attribute classifiers to describe images. Since then, a large amount of end-to-end IC models have been developed, eliminating the need for external object detector \cite{anderson2018bottom,vinyals2015show,herdade2019image,li2019entangled,cornia2020meshed,shalev2021randomized,ji2021improving}. However, several works \cite{agrawal2019nocaps,lu2018neural,tran2016rich} also leveraged the output of a pre-trained object detector to tackle the task of \textit{novel object description}. In this task, the IC model is given an image containing an unknown object, and a caption describing the object with the presented scene should be generated. Most commonly, external knowledge from a pre-trained object detector is facilitated for describing the unknown objects.

The task of novel object description overlaps with the task of OOD detection in the sense that both tasks need to deal with unknown objects. However, OOD image definition is not limited to images containing unknown objects but rather to a broad range of types such as corrupted and low-quality images or even images containing random noise. Moreover, in OOD detection, the model is expected to express its uncertainty in the caption to allow the rejection of the instance. In contrast, in novel object description, the generated caption is expected to include the unknown object.

While the proposed methods for novel object description gained success at describing images containing unknown objects, they mainly rely on the output of an object detector. Detecting whether an image is OOD \cite{hendrycks2016baseline} or contains novel objects \cite{agrawal2019nocaps} based on the outputs of the object detector is not sufficient for real-world applications since the number of classes that can be detected by the object detector is finite and limited, even when trained on massive image classification datasets. Additionally, object detectors include in their mechanism a classification module which by itself is vulnerable to OOD examples \cite{hendrycks2016baseline}, causing the entire process to be error-prone to both classification and detection errors made by the external detection model. 
Therefore, we find the task of OOD detection in IC an important complementary task to the novel object detection task.
\section{Notations and Definitions}\label{notations}
IC is the task of generating a natural language sentence
that accurately describes a given image.
Neural generative IC models typically consist of a CNN or visual Transformer, which is responsible for encoding the input image, and an RNN or Transformer that decodes descriptive sentences word-by-word conditioned on the image encoding. 

Commonly, the models trained as follows: assume a training dataset, $D_{train}=\{(I_i, S_i)\}^N_{i=1}$, composed of image-sentence pairs, where $I_i$ is the \textit{i}-th image and $S_i$ is the corresponding descriptive sentence. 
The parameters of the model, denoted by $\theta$, are found by MLE, which directly maximizes the log-likelihood of
the ground-truth description given the image, formally:
\begin{equation}\label{argmax_train}
    \theta^* = \argmax_{\theta}\sum_{(S_i,I_i)\in D_{train}} \log P(S_i|I_i;\theta)
\end{equation}
Since $S_i$ composed of a sequence of words $S_{i1}, ...,S_{ij}$, where $j$ is the length of $S_i$, a chain rule is applied to model the joint probability over the sentence words (we omit $\theta$ for simplicity):
\begin{equation*}
    \log P(S_i|I_i) = \sum_{z=1}^{z=j} \log P(S_{iz}|I, S_{i1}, ..., S_{iz-1})
\end{equation*}
This optimization requires assigning the highest probability to the ground-truth sentence, which is the sum of the log probabilities of its composing words.

At inference, a decoding strategy is applied, searching for the best describing sentence.
A \textit{greedy} decoding is the one that, at each decoding step, chooses the most probable next word. Another widely spread heuristic decoding strategy is \textit{Beam-search-K (BS-K)}, which maintains the $K$ most probable partial sequences until each decoding step. Notice that the greedy decoding is a special case of BS-K, where $K=1$. The BS-K does not guarantee finding the most probable sentence and leads to degenerated sentences when $K$ is large \cite{holtzman2019curious}. Several sampling-based decoding algorithms have been proposed. \textit{Top-K} decoding \cite{fan2018hierarchical}, is a method that samples at each timestamp a word from the K most probable next words. Later, \textit{Nucleus sampling (NS-p)} decoding strategy has been proposed \cite{holtzman2019curious}. NS-p suggests sampling from the top p portion of the probability mass instead of relying on a fixed top K candidate pool. 

\section{Problem Formulation}
In this paper, we are interested in the problem of out-of-distribution detection in IC: can we detect whether the given image is from a different distribution than the training data? Does the generated caption relate to the OOD image?

Denote by $\mathcal{S}$ the infinite set of all possible sentences and by $\mathcal{I}$ the infinite set of all possible images. Assume each training example, $(S_i,I_i)\in\mathcal{S} \times \mathcal{I}$, is drawn from a fixed but unknown distribution $\rho$. In OOD detection, a model is expected to perform well on unseen images drawn from the same distribution $\rho$ (\textit{in-distribution} images) and identify images drawn from a different distribution, $\mu$, to prevent the generation of unrelated captions.
Since we train the model to maximize the probability of the ground-truth captions for images drawn from $\rho$, its behavior on images drawn from $\mu$ is unexpected. Therefore, we are interested in exploring whether MLE training allows models to capture the relatedness between the generated caption and the given image. If the model learned to factor in the relatedness, it would be reasonable to assume that in-distribution images will result in higher likelihood captions than the OOD images. Meaning:
\begin{equation}
    \log P(S_{I_\rho}|I_\rho) > \log P(S_{I_\mu}|I_\mu)
\end{equation}
Where $I_{\rho}$, $I_{\mu}$ are in- and out- of-distribution images, respectively, and $S_{I_{\rho}}$, $S_{I_\mu}$ are the corresponding generated captions.

 In- and out-of-distribution images are not a strict dichotomy. There are various types of OOD images, such as images containing unknown objects (that are not appearing in the training set) or random noise. Identifying images containing unknown objects, which the model cannot accurately describe, is more challenging than detecting random noise images due to their relative visual closeness to in-distribution images.

We find that OOD images can be effectively detected using the generated captions' likelihood score. This finding implies that even though MLE training maximizes the sentence likelihood for in-distribution images only, it still allows the model to express its uncertainty when given OOD images by generating low probability captions. On the application side, the generated captions for OOD instances can be rejected based on the captions' likelihood to prevent the output of irrelevant captions.
\begin{figure}
  \centering
  \centerline{\includegraphics[width=9cm]{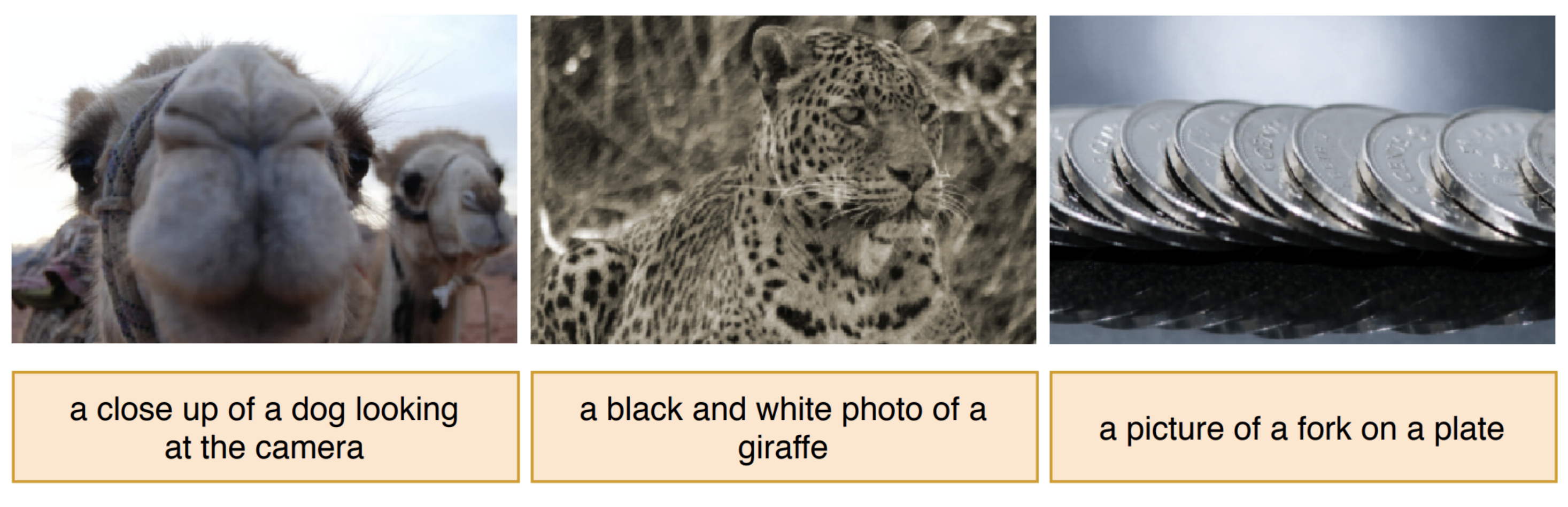}}
  \caption{Images from the cropped unknown object set
with captions generated by Top-Down model.}
  \label{fig:samples}
\vspace{-.3cm}
\end{figure}

\section{Experiments}
\begin{figure*}[t]
  \centering
  \centerline{\includegraphics[width=17cm]{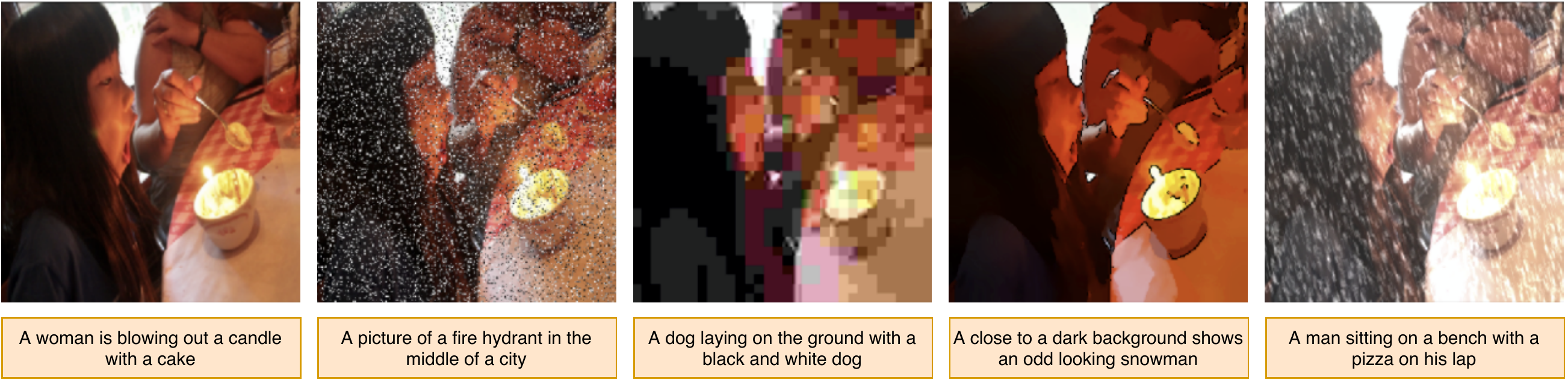}}
  \caption{Comparison between captions generated for (left to right): in-distribution image, and out-of-distribution corrupted images using: salt-and-pepper noise, JPEG compression, cartoon corruption, and snowflakes corruption}
  \label{fig:pert_img}
\end{figure*}
In this section, we demonstrate the effectiveness of the likelihood score at OOD detection by evaluating the performance of six widespread IC models. We also train an image classification model and compare its performance by applying two OOD detection methods. We start by describing the models in section \ref{baseline_section}. Then, in section \ref{ood_set_section} we describe the experimental setup including the OOD datasets. Evaluation metrics are described in \ref{evaluation_metrics}. Lastly, in section \ref{result_section}, we show the detection results and analyze the generated captions of both in- and out- of-distribution images.

\subsection{Baseline and Evaluated Models}\label{baseline_section}
As our captioning models, we trained two of the most widespread RNN-based and four Transformer-based IC models. RNN-based models: (1) Top-Down \cite{anderson2018bottom} (2) Show-Tell \cite{vinyals2015show}. We set a pre-trained ResNet-101 as the image encoder component in both models. 
Transformer models: (1) ORT \cite{herdade2019image} (2) ETA \cite{li2019entangled} (3) M2 \cite{cornia2020meshed} (4) GET \cite{ji2021improving}. 

To compare the performance of the IC models with OOD detection methods based on image-classification models, we trained a classification baseline model (ResNet-101) on objects appearing in the training datasets (we list the labels in the appendix). For the classification model we use two popular OOD detection methods: (1) Max probability score (MSP) \cite{hendrycks2016baseline}  (2) Generalized ODIN (GODIN)\cite{hsu2020generalized}. Notice that the classification and the captioning models are not exposed to OOD objects during training for a fair comparison.

For the captioning models, since the decoding strategies search for the most probable sentence differently, we evaluate captions generated by several decoding strategies to examine the effectiveness of the likelihood score at OOD detection. The strategies we compare are Greedy, BS-10, Top-10, and NS-0.8.

We train and evaluate each of the models on -  COCO \cite{lin2014microsoft} and Flickr-8K \cite{hodosh2013framing} datasets.

\subsection{Experimental Setup}
\label{ood_set_section}
For evaluating OOD detection in IC, we adopt the popular experimental settings of OOD detection in image \cite{hendrycks2016baseline} and text \cite{hendrycks2020pretrained} classification. In the proposed setup, models are required to distinguish between a set of in-distribution samples and multiple OOD sets in the following setup: First, a model is trained over a training dataset $D^{IN}_{train}$ (without the exposure to OOD instances). Afterward, the model is evaluated over the test set $D^{IN}_{test}$, and the predicted probability for each sample is recorded. Then, to assess the detection ability, the model is evaluated over several OOD datasets, denoted as $D^{OUT}_{1}$, ..., $D^{OUT}_{p}$, and the predicted probability for each sample in the sets is recorded. Lastly, a measure of the separation between the recorded probabilities for $D^{IN}_{test}$ and each of the OOD sets $D^{OUT}_{i}$ is calculated.

We follow this setup and conduct several experiments wherein we create two groups. The first group contains the likelihood scores of captions generated for in-distribution images. This group is shared across all the experiments. The second group contains the likelihood scores of captions generated for OOD images. We consider eight types of OOD image sets in various degrees of difficulty, described later in this section. We compare each of the groups created from the OOD sets with the group created from the in-distribution set and measure the separation between the two using the evaluation metrics described in section \ref{evaluation_metrics}. 
An IC model with a perfect OOD detection ability will allow setting a fixed threshold of $T$, where all captions generated for in-distribution images will have higher likelihood scores, and captions for OOD images will be assigned with lower scores. 

For evaluating the OOD detection methods by the classification model, we follow the same setup but consider the predicted model's confidence as described in the original papers.

\subsubsection{Out-of-Distribution Image Sets}
We evaluate the performance on eight different OOD image sets. Two of the sets composed of images with no object appearing in them. Additional two sets composed of images containing unknown objects (samples can be seen in Fig-\ref{fig:ood_samples} and Fig-\ref{fig:samples}). The last four sets are composed of algorithmically corrupted in-distribution images. The details of the sets are as follows:
    
     \textbf{Unknown objects:} For creating this set we collected images containing a scene with objects that are not appearing the training sets. The images are sourced from the \textit{Open Images V4} \cite{krasin2017openimages}, which is a publicly available human-annotated object detection dataset, containing 14.6M bounding boxes of 600 objects in \textasciitilde{$1.7M$} images. From this set, we manually collected 38 objects which are not appearing in the COCO set and constructed a set of 4067 images containing unknown objects in natural scenes. The full list of the extracted objects can be found in the appendix.
    
    \textbf{Cropped unknown objects:} 
     Since Open images V4 composed of natural images where the unknown objects are appearing with a scene in the background, we isolate the objects from the presented scene and create a set of images containing a single object only. For creating this set, we consider images from the \textit{Unknown objects} set, and crop the unknown objects using the corresponding ground-truth bounding boxes. Examples can be
seen in Fig-\ref{fig:samples}
    
    \textbf{Google's cartoon set:} A publicly available dataset of 10K 2D cartoon avatar images\footnote{https://google.github.io/cartoonset/}. The cartoons vary in 10 artwork categories, 4 color categories, and 4 proportion categories, with a total of \textasciitilde{$10^{13}$} possible combinations.
    
    \textbf{Random noise:} We generate 5K random noise images by randomly drawing each pixel's value. 
    
    \textbf{Corrupted image sets:}
    Since COCO and Flickr training sets contain high-quality images, we apply on the corresponding test sets four types of algorithmically generated corruptions to create the following OOD sets: (1) \textbf{JPEG corruption}: consisting of images corrupted by JPEG compression. (2) \textbf{Salt-and-pepper corruption}: consisting of images added with salt-and-pepper noise. (3) \textbf{Snow corruption}: consisting of images added with white noise, imitating images taken in snowy weather (4) \textbf{Cartoon corruption}: consisting of images converted to cartoon style. In  Fig-\ref{fig:pert_img} we demonstrate the considered corruptions.
     
    These corruptions are a subset of the corruptions used to benchmark models' robustness to common corruptions and perturbations \cite{hendrycks2019benchmarking, shalev2020redesigning}.  The corruptions are generated using ImgAug \cite{imgaug} library. We normalize the OOD images using the same normalization parameters used for normalizing the training set.

\begin{figure}[t]
  \centering
  \centerline{\includegraphics[width=8cm]{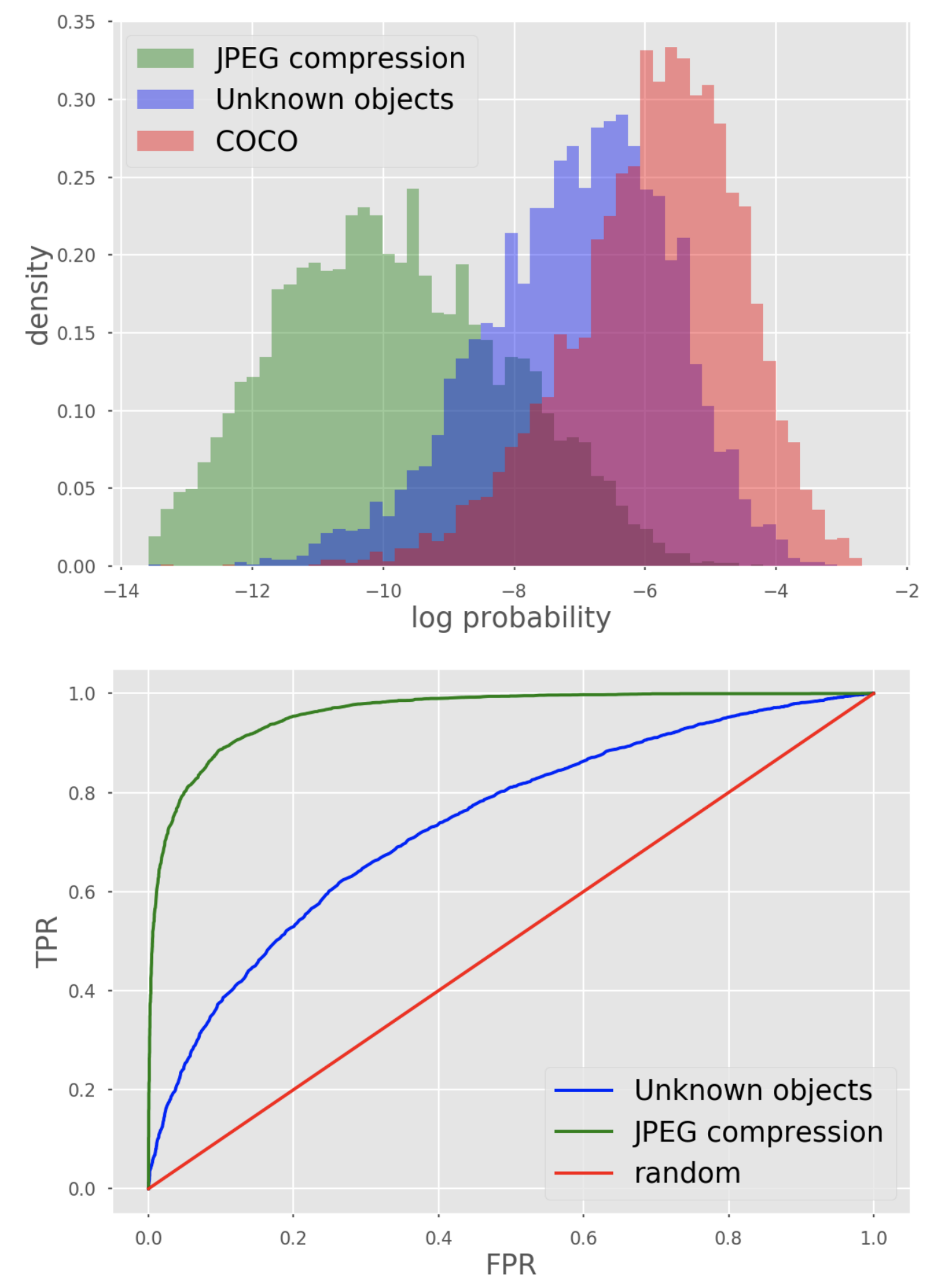}}
  \caption{The upper figure compares the log-likelihood scores of captions generated for in- and out-of-distribution images. The lower figure shows the ROC plot for the distributions above and compares it against a random detector. The captions are produced by the Top-Down model trained on the COCO dataset.}
  \label{fig:decoding_dif}
\end{figure}
\begin{table*}[t!]

\centering
\caption{IC evaluation metrics results for models trained on COCO dataset. We evaluate each metric on (from left to right): “Karpathy” test split (in bold), JPEG compressed set, salt-and-pepper set, snowflakes set ,and cartoon-corruption set.}
\begin{tabular}{lccc}
\toprule
 Model  & Cider & BLEU-4 & ROUGE-L \\
\midrule
\midrule
 Top-Down&  \textbf{1.16}\textbf{/} 0.09\textbf{/} 0.18\textbf{/} 0.41\textbf{/} 0.33 & \textbf{0.35}\textbf{/} 0.05\textbf{/} 0.09\textbf{/} 0.17\textbf{/} 0.13 & \textbf{0.53}\textbf{/} 0.29\textbf{/} 0.33\textbf{/} 0.36\textbf{/} 0.37      \\

 \midrule
 Show-Tell&  \textbf{0.98}\textbf{/} 0.17\textbf{/} 0.19\textbf{/} 0.36\textbf{/} 0.37 & \textbf{0.31}\textbf{/} 0.07\textbf{/} 0.06\textbf{/} 0.13\textbf{/} 0.14 & \textbf{0.51}\textbf{/} 0.30\textbf{/} 0.31\textbf{/} 0.38\textbf{/} 0.39      \\
   \midrule
 ORT&  \textbf{1.28}\textbf{/} 0.08\textbf{/} 0.16\textbf{/} 0.42\textbf{/} 0.38 & \textbf{0.38}\textbf{/} 0.06\textbf{/} 0.06\textbf{/} 0.11\textbf{/} 0.12 & \textbf{0.58}\textbf{/} 0.27\textbf{/} 0.31\textbf{/} 0.36\textbf{/} 0.37      \\
 \midrule
  ETA&  \textbf{1.26}\textbf{/} 0.13\textbf{/} 0.17\textbf{/} 0.39\textbf{/} 0.38 & \textbf{0.38}\textbf{/} 0.06\textbf{/} 0.06\textbf{/} 0.12\textbf{/} 0.13 & \textbf{0.58}\textbf{/} 0.29\textbf{/} 0.32\textbf{/} 0.36\textbf{/} 0.37      \\
   \midrule
  M2&  \textbf{1.30}\textbf{/} 0.14\textbf{/} 0.18\textbf{/} 0.38\textbf{/} 0.37 & \textbf{0.38}\textbf{/} 0.07\textbf{/} 0.08\textbf{/} 0.13\textbf{/} 0.13 & \textbf{0.58}\textbf{/} 0.30\textbf{/} 0.31\textbf{/} 0.36\textbf{/} 0.37      \\
   \midrule
  GET&  \textbf{1.30}\textbf{/} 0.14\textbf{/} 0.19\textbf{/} 0.39\textbf{/} 0.38 & \textbf{0.38}\textbf{/} 0.08\textbf{/} 0.06\textbf{/} 0.12\textbf{/} 0.13 & \textbf{0.58}\textbf{/} 0.30\textbf{/} 0.31\textbf{/} 0.36\textbf{/} 0.38      \\
\bottomrule
\end{tabular}
\label{ic_metrics}

\vspace{-.3cm}
\end{table*}

\subsection{Evaluation Metrics}\label{evaluation_metrics}
For evaluating the detection rate, we compare the likelihood scores of the generated sentences for the in-distribution image with the scores of the generated sentences for each of the OOD sets and measure how well they can be separated.
For measuring the separation, we use the following metrics:

     \textbf{Area under the Receiver Operating Characteristic curve (AUROC):} The metric was adopted by \cite{hendrycks2016baseline} for OOD detection in image classification task. The ROC curve depicts the relationship between \textit{true positive rate} and \textit{false positive rate}. The metric is a threshold-independent performance evaluation that can be interpreted as the probability that a positive example has a greater detector score/value than a negative example. A perfect detector corresponds to AUROC of 1. 
     
      \textbf{Area under the Precision-Recall curve:} As the AUROC is not ideal when the positive and negative classes have greatly differing base rates, another threshold independent metric was proposed for OOD detection \cite{hendrycks2016baseline}. The PR curve is a graph showing the \textit{precision} and \textit{recall} against each other. The metrics PRin and PRout in Tables denote the area under the precision-recall curve where in- and out- of-distribution images are specified as positives, respectively. A perfect detector has an AUPR of 1.
      
     \textbf{Bhattacharyya distance:} The distance metric was introduced by \cite{bhattacharyya1943measure} and later adopted by \cite{reyes2006bhattacharyya} for feature selection. Bhattacharyya Distance (BD) measures the similarity of two discrete or continuous distributions. 

    We use this metric to measure the amount of overlap between the distributions of the likelihood scores of caption generated for in- and out-of-distribution images. Higher BD value means that the overlap between the distributions is smaller. Hence, perfect OOD detection will result in $BD=\infty$.

\begin{table}[h]
\caption{Results for in- and out-of-distribution detection over the aggregation of the OOD sets. (COCO as an in-distribution set)}
\centering
\begin{tabular}{lrrrr}
\toprule
Model  & ROC$\uparrow$ & PRin$\uparrow$ & PRout$\uparrow$ & BD$\uparrow$ \\
\midrule
\midrule
        Top-Down  &  0.84 & 0.87 & 0.85 & 0.45      \\
        \midrule
         Show-Tell & 0.82 & 0.82 & 0.84 & 0.39      \\
\midrule
        ORT & 0.90 & 0.89 & 0.91 & 0.51      \\
\midrule
        ETA & 0.89 & 0.89 & 0.90 & 0.50      \\
\midrule
        M2 & 0.88 & 0.87 & 0.90 & 0.49      \\
\midrule
        GET & 0.90 & 0.90 & 0.90 & 0.53      \\
\midrule
\midrule
         MSP   & 0.73 & 0.72 & 0.74 & 0.10      \\
\midrule
         GODIN   & 0.76 & 0.74 & 0.78 & 0.18      \\         
                      
\bottomrule
\end{tabular}
\label{tab:ood_results_sets_aggregate}
\end{table}

\begin{table}[h]
\caption{Results for in- and out-of-distribution detection over the aggregation of the OOD sets. (Flickr-8K as an in-distribution set)}
\centering
\begin{tabular}{lrrrr}
\toprule
Model  & ROC$\uparrow$ & PRin$\uparrow$ & PRout$\uparrow$ & BD$\uparrow$ \\
\midrule
\midrule
        Top-Down  &  0.82 & 0.81 & 0.84 & 0.41      \\
\midrule
         Show-Tell & 0.81 & 0.80 & 0.81 & 0.36      \\
\midrule
        ORT & 0.89 & 0.88 & 0.90 & 0.50      \\
\midrule
        ETA & 0.87 & 0.88 & 0.86 & 0.49     \\
\midrule
        M2 & 0.88 & 0.87 & 0.89 & 0.51      \\
\midrule
        GET & 0.88 & 0.87 & 0.89 & 0.51      \\
\midrule
\midrule
         MSP   & 0.74 & 0.73 & 0.75 & 0.19      \\
\midrule
         GODIN   & 0.78 & 0.77 & 0.78 & 0.23      \\        
                      
\bottomrule
\end{tabular}
\label{tab:append_ood_results_sets_aggregate}
\end{table}
\subsection{Results}\label{result_section}
We start by aggregating all the OOD image sets and evaluating the detection rate. Tables \ref{tab:ood_results_sets_aggregate} and \ref{tab:append_ood_results_sets_aggregate} summarize the results for models trained on COCO and Flickr-8k datasets, respectively. As results indicate, the generated captions' likelihood scores can detect OOD instances significantly better than the methods based on image-classification models. Since the captions generated for the OOD images are assigned with a lower likelihood than unseen in-distribution images, they can be rejected with high rates by thresholding the likelihood score. Meaning that instead of applying two separate models where the first decide whether an image is OOD or not, and the second is the IC model responsible for generating a relevant caption, a single IC model can be applied. We also notice that Transformer-based IC models can detect OOD instances with higher rates than RNN-based models. These results are aligned with \cite{hendrycks2020pretrained} that showed that Transformer-based models used for \emph{text classification} are better OOD detectors. Next, we analyze the detection of each OOD set independently. We start by examining the corrupted image sets, and afterward, we analyze the unknown object, random noise, and cartoon image sets.

\textbf{Corrupted image sets:}
recall that these sets consist of corrupted in-distribution images taken from the corresponding test sets. To demonstrate the severity of the corruptions, we generate captions for each of the corrupted sets and measure the standard IC evaluation metrics against the corresponding ground-truth captions. Results, summarized in Table-\ref{ic_metrics}, show a significant drop in all metrics when models are evaluated over the corrupted sets since the generated captions are irrelevant to the given images. For comparison, randomly generated captions (created by randomly drawing tokens from the vocabulary without considering the image) achieve the following scores: a Cider of 0.06, BLEU-4 of 0.04, and ROUGE-L of 0.24. 
However, despite the severe drop caused by these sets, we find that the corrupted images can be effectively detected by considering the captions' likelihood scores. The detection results of the models trained on the COCO dataset appear in Table-\ref{tab:ood_rnn_baseline} (results for Flickr data sets are consistent). 

In Fig-\ref{fig:decoding_dif}, we demonstrate the likelihood of the JPEG compression set against COCO's test set. The ROC curve for the JPEG compression set shows that when 80\% of the compressed images are rejected, less than 3\% of in-distribution images are falsely rejected (at the point where the TPR and FPR are 97\%, 20\% respectively). For comparison, a random detector that randomly assigns a score for a caption would reject only 3\% of the OOD images while falsely rejecting 3\% of the in-distribution images (at the point where the TPR and FPR are both 97\%).

\begin{table}[h]
\caption{Results for in- and out-of-distribution detection over the corrupted OOD image sets.}
\centering
\small
\begin{tabular}{llrrrr}
\toprule
Model &OOD set   & ROC$\uparrow$ & PRin$\uparrow$ & PRout$\uparrow$ & BD$\uparrow$ \\
\midrule
\midrule
    \multirow{4}{0cm}{ORT} 

&Salt-and-pepper& 0.924 & 0.915 & 0.918 &    0.523   \\\cmidrule{2-6}
&JPEG compression& 0.931 & 0.937 & 0.929 &    0.503   \\\cmidrule{2-6}
&Snow flakes& 0.829 & 0.801 & 0.842 &    0.184   \\\cmidrule{2-6}
&Cartoon corruption& 0.836 & 0.822 & 0.851 &    0.210   \\
\midrule
    \multirow{4}{0cm}{ETA}
&Salt-and-pepper& 0.909 & 0.922 & 0.900 &    0.511   \\\cmidrule{2-6}
&JPEG compression& 0.939 & 0.944 & 0.923 &    0.555   \\\cmidrule{2-6}
&Snow flakes& 0.801 & 0.789 & 0.832 &    0.134   \\\cmidrule{2-6}
&Cartoon corruption& 0.833 & 0.813 & 0.836 &    0.202   \\
\midrule
    \multirow{4}{0cm}{Top-Down} 
&Salt-and-pepper& 0.911 & 0.910 & 0.909 &    0.455   \\\cmidrule{2-6}
&JPEG compression& 0.961 & 0.962 & 0.968 &    0.797   \\\cmidrule{2-6}
&Snow flakes& 0.778 & 0.739 & 0.791 &    0.111   \\\cmidrule{2-6}
&Cartoon corruption& 0.803 & 0.785 & 0.809 &    0.149   \\
\midrule
    \multirow{4}{0cm}{Show-Tell} 
&Salt-and-pepper& 0.863 & 0.849 & 0.851 &    0.292   \\\cmidrule{2-6}
&JPEG compression& 0.870 & 0.868 & 0.851 &    0.321   \\\cmidrule{2-6}
&Snow flakes& 0.752 & 0.693 & 0.772 &    0.103   \\\cmidrule{2-6}
&Cartoon corruption& 0.725 & 0.682 & 0.712 &    0.053   \\
\midrule
\midrule
    \multirow{4}{0cm}{MSP} &Salt-and-pepper  &  0.891  & 0.901  & 0.854  &  0.498      \\\cmidrule{2-6}

&JPEG compression& 0.822 & 0.854  & 0.812  & 0.349\\\cmidrule{2-6}

&Snow flakes &  0.712 & 0.735  & 0.701 & 0.118        \\\cmidrule{2-6}
&Cartoon corruption& 0.852 & 0.856 & 0.829 &    0.219   \\
\midrule
    \multirow{4}{0cm}{GODIN}&Salt-and-pepper  &  0.843  & 0.810  & 0.866  &  0.432      \\\cmidrule{2-6}

&JPEG compression& 0.849 & 0.841  & 0.850  & 0.422        \\\cmidrule{2-6}
&Snow flakes &  0.739 & 0.726  & 0.744 & 0.127        \\\cmidrule{2-6}
&Cartoon corruption& 0.741 & 0.765 & 0.704 &    0.089   \\
\bottomrule
\bottomrule
\end{tabular}
\label{tab:ood_rnn_baseline}
\end{table}

\begin{table}[h]
\centering
\caption{Results for in- and out-of-distribution detection over the non-corrupted OOD image sets.}
\small
\begin{tabular}{llrrrr}
\toprule
Model &OOD set   & ROC$\uparrow$ & PRin$\uparrow$ & PRout$\uparrow$ & BD$\uparrow$ \\
\midrule
\midrule
    \multirow{4}{0cm}{ORT} &Unknown objects  &  0.769  & 0.771  & 0.763  &  0.132      \\\cmidrule{2-6}

&Random noise& 0.985 & 0.971  & 0.992  & 1.691        \\\cmidrule{2-6}

&Google's cartoon set &  0.986 & 0.988  & 0.979 & 1.198        \\\cmidrule{2-6}
&Cropped unknown objects& 0.767 & 0.764 & 0.782 &    0.131   \\

\midrule
    \multirow{4}{0cm}{ETA}&Unknown objects  &  0.755  & 0.764  & 0.701  &  0.119      \\\cmidrule{2-6}

&Random noise& 0.981 & 0.976  & 0.984  & 1.531        \\\cmidrule{2-6}
&Google's cartoon set &  0.978 & 0.969  & 0.984 & 1.101        \\\cmidrule{2-6}
&Cropped unknown objects& 0.781 & 0.771 & 0.793 &    0.163   \\
\midrule
    \multirow{4}{0cm}{Top-Down} &Unknown objects  &  0.736  & 0.737  & 0.716  &  0.115      \\\cmidrule{2-6}

&Random noise& 0.965 & 0.981  & 0.899  & 1.682        \\\cmidrule{2-6}

&Google's cartoon set &  0.980 & 0.985  & 0.964 & 1.181        \\\cmidrule{2-6}
&Cropped unknown objects& 0.759 & 0.762 & 0.732 &    0.128   \\
\midrule
    \multirow{4}{0cm}{Show-Tell}&Unknown objects  &  0.727  & 0.722  & 0.704  &  0.097      \\\cmidrule{2-6}

&Random noise& 0.954 & 0.969  & 0.921  & 0.995        \\\cmidrule{2-6}
&Google's cartoon set &  0.930 & 0.943  & 0.910 & 0.595        \\\cmidrule{2-6}
&Cropped unknown objects& 0.722 & 0.721 & 0.699 &    0.092   \\
\midrule
\midrule
    \multirow{4}{0cm}{MSP} &Unknown objects  &  0.641  & 0.656  & 0.629  &  0.099      \\\cmidrule{2-6}

&Random noise& 0.899 & 0.903  & 0.886  & 0.349\\\cmidrule{2-6}

&Google's cartoon set &  0.912 & 0.922  & 0.909 & 0.498        \\\cmidrule{2-6}
&Cropped unknown objects& 0.702 & 0.689 & 0.711 &    0.094   \\
\midrule
    \multirow{4}{0cm}{GODIN}&Unknown objects  &  0.706  & 0.699  & 0.711  &  0.102      \\\cmidrule{2-6}

&Random noise& 0.969 & 0.984  & 0.973  & 1.544        \\\cmidrule{2-6}
&Google's cartoon set &  0.963 & 0.977  & 0.965 & 1.001        \\\cmidrule{2-6}
&Cropped unknown objects& 0.786 & 0.782 & 0.789 &    0.171   \\
\bottomrule
\bottomrule
\end{tabular}
\label{tab:ood_ort_baseline}
\end{table}

\begin{figure}[t]
  \centering
  \centerline{\includegraphics[width=8cm]{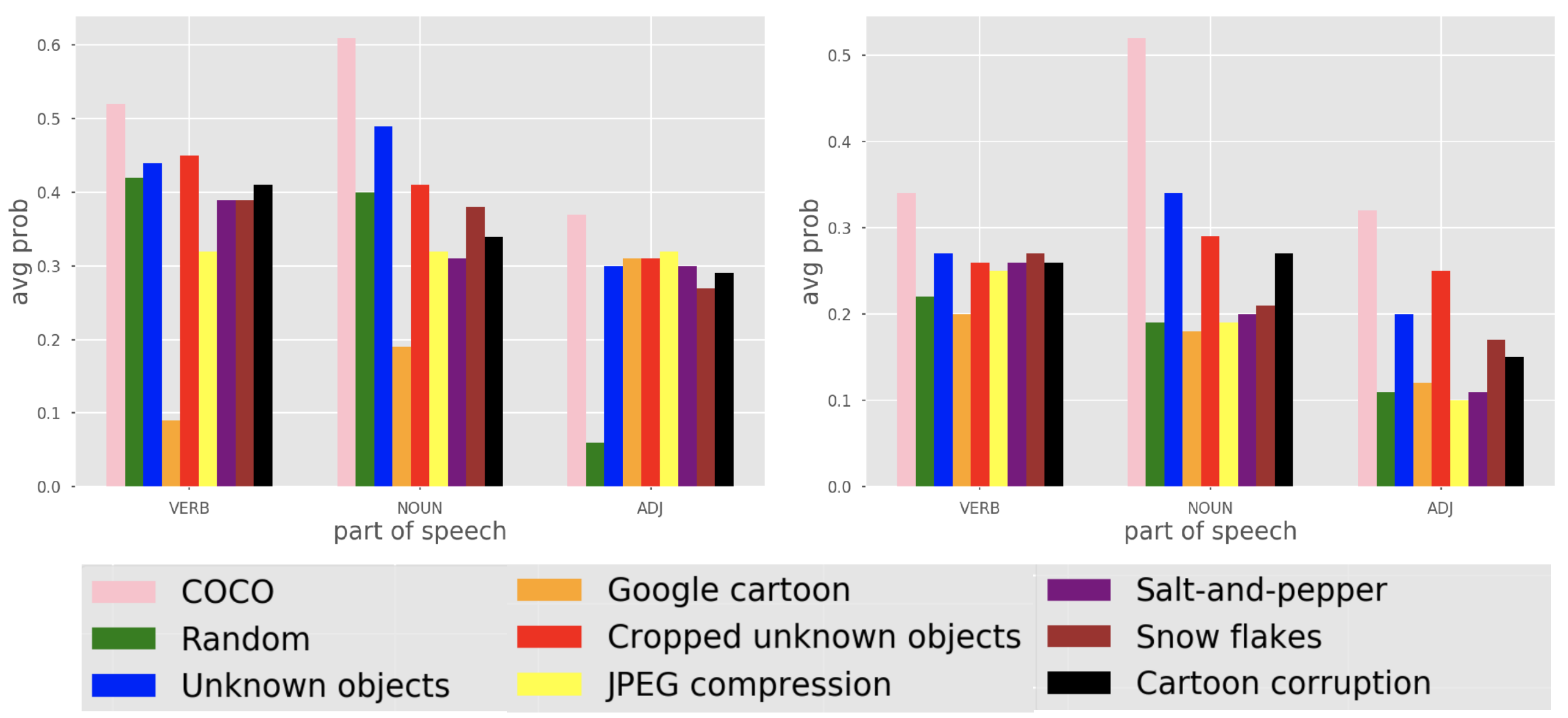}}
  \caption{Comparison between the average predicted probability for noun, verb, and adjective tokens.
  On the left panel we show the results of Show-Tell model with NS-0.8.
  On the right panel we show the results of Top-Down model with BS-10. }
  \label{fig:pos_diff}
\vspace{-.3cm}
\end{figure}

\textbf{Non-corrupted image sets:} moving forward to the next four OOD sets. These sets, composed of images containing unknown objects, cartoons, and noise, can also be detected at high rates using the likelihood score. Table-\ref{tab:ood_ort_baseline} summarizes the detection rates of the models. As can be seen, unsurprisingly, images containing random noise or cartoons are the easiest to detect, while images containing unknown objects are the hardest due to their visual closeness to in-distribution images.

Here as before, we find the Transformer-based models outperform. In all tested sets, the models outperform a random detector\footnote{AUROC is equal to 0.5, BD
is equal to 0, and AUPR is equal to the precision} and the classification-based models. To gain an intuition for why the classification models achieve lower results, we checked their confidence scores and found that they tend to be high even for objects that are not part of the training object (overconfidence when wrong). This was also observed in previous works \cite{guo2017calibration}.  

Recall that training the captioning models is done by MLE, which does not guarantee to produce less likely captions for OOD images. To better understand why exactly the generated captions for OOD images tend to have a lower likelihood, we extract the predicted probabilities of their visual counterpart tokens. More specifically, we consider the \textit{part-of-speech} of each token in the generated caption and focus on the predicted probabilities of \textit{noun, verb}, and \textit{adjective} tokens, as they should be visually grounded in the given image. Interestingly, in all models, the \textit{noun}, \textit{verb}, and \textit{adjective} tokens are predicted with lower probabilities for OOD images compared to  in-distribution images (see  Fig-\ref{fig:pos_diff}). In contrast, tokens from other \textit{part-of-speech} groups such as \textit{determiner} (a, an, the, etc.), \textit{adverbs} (very, there, where, etc.), and \textit{adposition} (in, to, during, etc.) are predicted with the same average probability for both in- and out -of-distribution images. Our intuition is that since the predicted visual tokens should match the image and fit the textual context, the model can learn the in-distribution from a richer signal compared to the classification model. We also noticed, in preliminary results, that as we increased the weight of the loss of visually grounded tokens during training (by multiplying the loss with a constant), the OOD detection rates increased while at the same time, the performance on in-distribution remained unchanged. We leave the exploration of this direction to future work.

\begin{table}[h]
\caption{Results for in- and out-of-distribution detection of Top-Down model using different decoding strategies.}
\centering
\small
\begin{tabular}{llrrrr}
\toprule
OOD set  & decoding & ROC$\uparrow$ & PRin$\uparrow$ & PRout$\uparrow$ & BD$\uparrow$ \\
\midrule
\midrule
Salt-and- & Greedy  & 0.831 & 0.842 & 0.809 & 0.230      \\
pepper                      & Top-10  & 0.597 & 0.614 & 0.562 & 0.025      \\
corruption                      & N-0.8  & 0.837 & 0.833 & 0.826 & 0.238      \\
\midrule
JPEG & Greedy  & 0.935 & 0.939 & 0.928 & 0.571      \\
compression                      & Top-10  & 0.657 & 0.690 & 0.604 & 0.061      \\
corruption                      & N-0.8  & 0.892 & 0.897 & 0.877 & 0.379      \\
\midrule
Snow & Greedy  & 0.706 & 0.678 & 0.703 & 0.091      \\
flakes                      & Top-10  & 0.603 & 0.610 & 0.574 & 0.013      \\
corruption                      & N-0.8  & 0.696 & 0.675 & 0.685 & 0.076      \\
\midrule
Cartoon & Greedy  & 0.705 & 0.689 & 0.715 & 0.070      \\
corruption                      & Top-10  & 0.590 & 0.585 & 0.571 & 0.019      \\
                      & N-0.8  & 0.757 & 0.736 & 0.755 & 0.105      \\
\midrule
\midrule
    Unknown     & Greedy  & 0.691 & 0.690 & 0.670 & 0.068      \\
    objects                  & Top-10  & 0.555 & 0.566 & 0.531 & 0.013      \\
                      & N-0.8  & 0.676 & 0.669 & 0.659 & 0.057      \\
\midrule
Random & Greedy  & 0.809 & 0.863 & 0.766 & 0.720      \\
noise                      & Top-10  & 0.657 & 0.685 & 0.610 & 0.058      \\
                      & N-0.8  & 0.911 & 0.924 & 0.891 & 0.470      \\
\midrule
Google's      & Greedy  & 0.892 & 0.922 & 0.842 & 0.515      \\
cartoon                      & Top-10  & 0.566 & 0.609 & 0.531 & 0.028      \\
set                     & N-0.8  & 0.865 & 0.884 & 0.830 & 0.336      \\
\midrule
Cropped & Greedy  & 0.708 & 0.713 & 0.679 & 0.082      \\
unknown                      & Top-10  & 0.549 & 0.559 & 0.525 & 0.012      \\
objects                      & N-0.8  & 0.760 & 0.761 & 0.734 & 0.130    \\

\bottomrule
\end{tabular}
\label{tab:ood_results_corrupted}
\vspace{-.5cm}
\end{table}

\begin{table}[h]
\caption{Results for in- and out-of-distribution detection of Show-Tell model using different decoding strategies.}
\centering
\small
\begin{tabular}{llrrrr}
\toprule
OOD set  & decoding & ROC$\uparrow$ & PRin$\uparrow$ & PRout$\uparrow$ & BD$\uparrow$ \\
\midrule
\midrule
Salt-and- & Greedy  & 0.755 & 0.761 & 0.721 & 0.139      \\
pepper                      & Top-10  & 0.561 & 0.584 & 0.537 & 0.017      \\
corruption                     & N-0.8  & 0.682 & 0.688 & 0.656 & 0.064      \\
\midrule
JPEG & Greedy  & 0.761 & 0.777 & 0.742 & 0.148      \\
compression                      & Top-10  & 0.635 & 0.649 & 0.607 & 0.042      \\
corruption                      & N-0.8  & 0.750 & 0.754 & 0.719 & 0.242      \\
\midrule
Snow & Greedy  & 0.698 & 0.666 & 0.687 & 0.087      \\
flakes                      & Top-10  & 0.586 & 0.588 & 0.571 & 0.012      \\
corruption                      & N-0.8  & 0.645 & 0.630 & 0.644 & 0.069      \\
\midrule
Cartoon & Greedy  & 0.676 & 0.653 & 0.674 & 0.044      \\
corruption                     & Top-10  & 0.588 & 0.591 & 0.572 & 0.011      \\
                      & N-0.8  & 0.672 & 0.653 & 0.661 & 0.038      \\
\midrule
\midrule
    Unknown     & Greedy  & 0.657 & 0.679 & 0.627 & 0.044      \\
    objects                  & Top-10  & 0.547 & 0.563 & 0.523 & 0.013      \\
                      & N-0.8  & 0.645 & 0.647 & 0.622 & 0.039      \\
\midrule
Random & Greedy  & 0.796 & 0.860 & 0.684 & 0.521      \\
noise                      & Top-10  & 0.630 & 0.657 & 0.590 & 0.038      \\
                      & N-0.8  & 0.815 & 0.851 & 0.809 & 0.310      \\
\midrule
Google's      & Greedy  & 0.863 & 0.877 & 0.843 & 0.333      \\
cartoon                      & Top-10  & 0.611 & 0.632 & 0.584 & 0.033      \\
set                      & N-0.8  & 0.836 & 0.853 & 0.814 & 0.257      \\
\midrule
Cropped & Greedy  & 0.649 & 0.667 & 0.617 & 0.049      \\
unknown                      & Top-10  & 0.548 & 0.564 & 0.531 & 0.013      \\
objects                      & N-0.8  & 0.641 & 0.642 & 0.616 & 0.041      \\
\bottomrule
\end{tabular}
\label{tab:ood_results_sets}
\end{table}

\textbf{Decoding strategies:}\label{decoding_strat}
 lastly, we investigate the influence of the decoding strategy on the OOD detection rates. We compare three additional types of decoding strategies: (1) greedy. (2) top-10 sampling. (3) nucleus sampling (N-0.8).
Results presented in Table-\ref{tab:ood_results_corrupted} and Table-\ref{tab:ood_results_sets} indicate that the influence of the decoding strategy has a high impact on the overall results and that beam search is the most effective for this task. We find Top-K as the least effective decoding method for this task, as it has higher chances of choosing low confidence tokens even for in-distribution samples. As some decoding strategies yield significantly superior results for OOD detection than others, we think that developing new techniques that consider the relatedness might boost the results further.

\section{Conclusions and Future Work}
To conclude, we formulated the task of OOD detection in image captioning and presented evaluation datasets and metrics. We demonstrated that IC models are able to express their uncertainty when facing OOD images by generating low probability captions. While the captions' likelihood is effective for detecting OOD images and is able to capture the image-caption relatedness, results indicate there is still room for improvement. We hope to inspire future work on developing new models, training methods, and decoding strategies for tackling the important task of \textit{"knowing"} when the model cannot describe the given image.

\bibliographystyle{ACM-Reference-Format}
\bibliography{sample-base}


\begin{thebibliography}{43}


\ifx \showCODEN    \undefined \def \showCODEN     #1{\unskip}     \fi
\ifx \showDOI      \undefined \def \showDOI       #1{#1}\fi
\ifx \showISBNx    \undefined \def \showISBNx     #1{\unskip}     \fi
\ifx \showISBNxiii \undefined \def \showISBNxiii  #1{\unskip}     \fi
\ifx \showISSN     \undefined \def \showISSN      #1{\unskip}     \fi
\ifx \showLCCN     \undefined \def \showLCCN      #1{\unskip}     \fi
\ifx \shownote     \undefined \def \shownote      #1{#1}          \fi
\ifx \showarticletitle \undefined \def \showarticletitle #1{#1}   \fi
\ifx \showURL      \undefined \def \showURL       {\relax}        \fi
\providecommand\bibfield[2]{#2}
\providecommand\bibinfo[2]{#2}
\providecommand\natexlab[1]{#1}
\providecommand\showeprint[2][]{arXiv:#2}

\bibitem[\protect\citeauthoryear{Agrawal, Desai, Wang, Chen, Jain, Johnson,
  Batra, Parikh, Lee, and Anderson}{Agrawal et~al\mbox{.}}{2019}]%
        {agrawal2019nocaps}
\bibfield{author}{\bibinfo{person}{Harsh Agrawal}, \bibinfo{person}{Karan
  Desai}, \bibinfo{person}{Yufei Wang}, \bibinfo{person}{Xinlei Chen},
  \bibinfo{person}{Rishabh Jain}, \bibinfo{person}{Mark Johnson},
  \bibinfo{person}{Dhruv Batra}, \bibinfo{person}{Devi Parikh},
  \bibinfo{person}{Stefan Lee}, {and} \bibinfo{person}{Peter Anderson}.}
  \bibinfo{year}{2019}\natexlab{}.
\newblock \showarticletitle{nocaps: novel object captioning at scale}. In
  \bibinfo{booktitle}{\emph{Proceedings of the IEEE International Conference on
  Computer Vision}}. \bibinfo{pages}{8948--8957}.
\newblock


\bibitem[\protect\citeauthoryear{Anderson, He, Buehler, Teney, Johnson, Gould,
  and Zhang}{Anderson et~al\mbox{.}}{2018}]%
        {anderson2018bottom}
\bibfield{author}{\bibinfo{person}{Peter Anderson}, \bibinfo{person}{Xiaodong
  He}, \bibinfo{person}{Chris Buehler}, \bibinfo{person}{Damien Teney},
  \bibinfo{person}{Mark Johnson}, \bibinfo{person}{Stephen Gould}, {and}
  \bibinfo{person}{Lei Zhang}.} \bibinfo{year}{2018}\natexlab{}.
\newblock \showarticletitle{Bottom-up and top-down attention for image
  captioning and visual question answering}. In
  \bibinfo{booktitle}{\emph{Proceedings of the IEEE Conference on Computer
  Vision and Pattern Recognition}}. \bibinfo{pages}{6077--6086}.
\newblock


\bibitem[\protect\citeauthoryear{Bhattacharyya}{Bhattacharyya}{1943}]%
        {bhattacharyya1943measure}
\bibfield{author}{\bibinfo{person}{Anil Bhattacharyya}.}
  \bibinfo{year}{1943}\natexlab{}.
\newblock \showarticletitle{On a measure of divergence between two statistical
  populations defined by their probability distributions}.
\newblock \bibinfo{journal}{\emph{Bull. Calcutta Math. Soc.}}
  \bibinfo{volume}{35} (\bibinfo{year}{1943}), \bibinfo{pages}{99--109}.
\newblock


\bibitem[\protect\citeauthoryear{Cornia, Stefanini, Baraldi, and
  Cucchiara}{Cornia et~al\mbox{.}}{2020}]%
        {cornia2020meshed}
\bibfield{author}{\bibinfo{person}{Marcella Cornia}, \bibinfo{person}{Matteo
  Stefanini}, \bibinfo{person}{Lorenzo Baraldi}, {and} \bibinfo{person}{Rita
  Cucchiara}.} \bibinfo{year}{2020}\natexlab{}.
\newblock \showarticletitle{Meshed-memory transformer for image captioning}. In
  \bibinfo{booktitle}{\emph{Proceedings of the IEEE/CVF Conference on Computer
  Vision and Pattern Recognition}}. \bibinfo{pages}{10578--10587}.
\newblock


\bibitem[\protect\citeauthoryear{Do, Duong, Dang, and Le}{Do
  et~al\mbox{.}}{2018}]%
        {do2018real}
\bibfield{author}{\bibinfo{person}{Truong-Dong Do}, \bibinfo{person}{Minh-Thien
  Duong}, \bibinfo{person}{Quoc-Vu Dang}, {and} \bibinfo{person}{My-Ha Le}.}
  \bibinfo{year}{2018}\natexlab{}.
\newblock \showarticletitle{Real-Time Self-Driving Car Navigation Using Deep
  Neural Network}. In \bibinfo{booktitle}{\emph{2018 4th International
  Conference on Green Technology and Sustainable Development (GTSD)}}. IEEE,
  \bibinfo{pages}{7--12}.
\newblock


\bibitem[\protect\citeauthoryear{Dognin, Melnyk, Mroueh, Padhi, Rigotti, Ross,
  Schiff, Young, and Belgodere}{Dognin et~al\mbox{.}}{2022}]%
        {dognin2020image}
\bibfield{author}{\bibinfo{person}{Pierre Dognin}, \bibinfo{person}{Igor
  Melnyk}, \bibinfo{person}{Youssef Mroueh}, \bibinfo{person}{Inkit Padhi},
  \bibinfo{person}{Mattia Rigotti}, \bibinfo{person}{Jarret Ross},
  \bibinfo{person}{Yair Schiff}, \bibinfo{person}{Richard~A Young}, {and}
  \bibinfo{person}{Brian Belgodere}.} \bibinfo{year}{2022}\natexlab{}.
\newblock \showarticletitle{Image captioning as an assistive technology:
  Lessons learned from vizwiz 2020 challenge}.
\newblock \bibinfo{journal}{\emph{Journal of Artificial Intelligence Research}}
  (\bibinfo{year}{2022}).
\newblock


\bibitem[\protect\citeauthoryear{Fan, Lewis, and Dauphin}{Fan
  et~al\mbox{.}}{2018}]%
        {fan2018hierarchical}
\bibfield{author}{\bibinfo{person}{Angela Fan}, \bibinfo{person}{Mike Lewis},
  {and} \bibinfo{person}{Yann Dauphin}.} \bibinfo{year}{2018}\natexlab{}.
\newblock \showarticletitle{Hierarchical Neural Story Generation}. In
  \bibinfo{booktitle}{\emph{Proceedings of the 56th Annual Meeting of the
  Association for Computational Linguistics (Volume 1: Long Papers)}}.
  \bibinfo{publisher}{Association for Computational Linguistics},
  \bibinfo{address}{Melbourne, Australia}, \bibinfo{pages}{889--898}.
\newblock
\urldef\tempurl%
\url{https://doi.org/10.18653/v1/P18-1082}
\showDOI{\tempurl}


\bibitem[\protect\citeauthoryear{Farhadi, Hejrati, Sadeghi, Young, Rashtchian,
  Hockenmaier, and Forsyth}{Farhadi et~al\mbox{.}}{2010}]%
        {farhadi2010every}
\bibfield{author}{\bibinfo{person}{Ali Farhadi}, \bibinfo{person}{Mohsen
  Hejrati}, \bibinfo{person}{Mohammad~Amin Sadeghi}, \bibinfo{person}{Peter
  Young}, \bibinfo{person}{Cyrus Rashtchian}, \bibinfo{person}{Julia
  Hockenmaier}, {and} \bibinfo{person}{David Forsyth}.}
  \bibinfo{year}{2010}\natexlab{}.
\newblock \showarticletitle{Every picture tells a story: Generating sentences
  from images}. In \bibinfo{booktitle}{\emph{European conference on computer
  vision}}. Springer, \bibinfo{pages}{15--29}.
\newblock


\bibitem[\protect\citeauthoryear{Granovsky, Shalev, Yacovzada, Frank, and
  Fine}{Granovsky et~al\mbox{.}}{2018}]%
        {granovsky2018actigraphy}
\bibfield{author}{\bibinfo{person}{Lena Granovsky}, \bibinfo{person}{Gabi
  Shalev}, \bibinfo{person}{Nancy Yacovzada}, \bibinfo{person}{Yotam Frank},
  {and} \bibinfo{person}{Shai Fine}.} \bibinfo{year}{2018}\natexlab{}.
\newblock \showarticletitle{Actigraphy-based sleep/wake pattern detection using
  convolutional neural networks}.
\newblock \bibinfo{journal}{\emph{arXiv preprint arXiv:1802.07945}}
  (\bibinfo{year}{2018}).
\newblock


\bibitem[\protect\citeauthoryear{Guo, Pleiss, Sun, and Weinberger}{Guo
  et~al\mbox{.}}{2017}]%
        {guo2017calibration}
\bibfield{author}{\bibinfo{person}{Chuan Guo}, \bibinfo{person}{Geoff Pleiss},
  \bibinfo{person}{Yu Sun}, {and} \bibinfo{person}{Kilian~Q Weinberger}.}
  \bibinfo{year}{2017}\natexlab{}.
\newblock \showarticletitle{On calibration of modern neural networks}. In
  \bibinfo{booktitle}{\emph{International Conference on Machine Learning}}.
  PMLR, \bibinfo{pages}{1321--1330}.
\newblock


\bibitem[\protect\citeauthoryear{Hendrycks and Dietterich}{Hendrycks and
  Dietterich}{2019}]%
        {hendrycks2019benchmarking}
\bibfield{author}{\bibinfo{person}{Dan Hendrycks} {and} \bibinfo{person}{Thomas
  Dietterich}.} \bibinfo{year}{2019}\natexlab{}.
\newblock \showarticletitle{Benchmarking Neural Network Robustness to Common
  Corruptions and Perturbations}.
\newblock \bibinfo{journal}{\emph{Proceedings of the International Conference
  on Learning Representations}} (\bibinfo{year}{2019}).
\newblock


\bibitem[\protect\citeauthoryear{Hendrycks and Gimpel}{Hendrycks and
  Gimpel}{2017}]%
        {hendrycks2016baseline}
\bibfield{author}{\bibinfo{person}{Dan Hendrycks} {and} \bibinfo{person}{Kevin
  Gimpel}.} \bibinfo{year}{2017}\natexlab{}.
\newblock \showarticletitle{A Baseline for Detecting Misclassified and
  Out-of-Distribution Examples in Neural Networks}.
\newblock \bibinfo{journal}{\emph{Proceedings of International Conference on
  Learning Representations}} (\bibinfo{year}{2017}).
\newblock


\bibitem[\protect\citeauthoryear{Hendrycks, Liu, Wallace, Dziedzic, Krishnan,
  and Song}{Hendrycks et~al\mbox{.}}{2020}]%
        {hendrycks2020pretrained}
\bibfield{author}{\bibinfo{person}{Dan Hendrycks}, \bibinfo{person}{Xiaoyuan
  Liu}, \bibinfo{person}{Eric Wallace}, \bibinfo{person}{Adam Dziedzic},
  \bibinfo{person}{Rishabh Krishnan}, {and} \bibinfo{person}{Dawn Song}.}
  \bibinfo{year}{2020}\natexlab{}.
\newblock \showarticletitle{Pretrained Transformers Improve Out-of-Distribution
  Robustness}. In \bibinfo{booktitle}{\emph{Proceedings of the 58th Annual
  Meeting of the Association for Computational Linguistics}}.
  \bibinfo{publisher}{Association for Computational Linguistics},
  \bibinfo{address}{Online}, \bibinfo{pages}{2744--2751}.
\newblock
\urldef\tempurl%
\url{https://doi.org/10.18653/v1/2020.acl-main.244}
\showDOI{\tempurl}


\bibitem[\protect\citeauthoryear{Herdade, Kappeler, Boakye, and Soares}{Herdade
  et~al\mbox{.}}{2019}]%
        {herdade2019image}
\bibfield{author}{\bibinfo{person}{Simao Herdade}, \bibinfo{person}{Armin
  Kappeler}, \bibinfo{person}{Kofi Boakye}, {and} \bibinfo{person}{Joao
  Soares}.} \bibinfo{year}{2019}\natexlab{}.
\newblock \showarticletitle{Image captioning: Transforming objects into words}.
\newblock \bibinfo{journal}{\emph{arXiv preprint arXiv:1906.05963}}
  (\bibinfo{year}{2019}).
\newblock


\bibitem[\protect\citeauthoryear{Hodosh, Young, and Hockenmaier}{Hodosh
  et~al\mbox{.}}{2013}]%
        {hodosh2013framing}
\bibfield{author}{\bibinfo{person}{Micah Hodosh}, \bibinfo{person}{Peter
  Young}, {and} \bibinfo{person}{Julia Hockenmaier}.}
  \bibinfo{year}{2013}\natexlab{}.
\newblock \showarticletitle{Framing image description as a ranking task: Data,
  models and evaluation metrics}.
\newblock \bibinfo{journal}{\emph{Journal of Artificial Intelligence Research}}
   \bibinfo{volume}{47} (\bibinfo{year}{2013}), \bibinfo{pages}{853--899}.
\newblock


\bibitem[\protect\citeauthoryear{Holtzman, Buys, Forbes, and Choi}{Holtzman
  et~al\mbox{.}}{2020}]%
        {holtzman2019curious}
\bibfield{author}{\bibinfo{person}{Ari Holtzman}, \bibinfo{person}{Jan Buys},
  \bibinfo{person}{Maxwell Forbes}, {and} \bibinfo{person}{Yejin Choi}.}
  \bibinfo{year}{2020}\natexlab{}.
\newblock \showarticletitle{The curious case of neural text degeneration}.
\newblock \bibinfo{journal}{\emph{Proceedings of International Conference on
  Learning Representations}} (\bibinfo{year}{2020}).
\newblock


\bibitem[\protect\citeauthoryear{Hsu, Shen, Jin, and Kira}{Hsu
  et~al\mbox{.}}{2020}]%
        {hsu2020generalized}
\bibfield{author}{\bibinfo{person}{Yen-Chang Hsu}, \bibinfo{person}{Yilin
  Shen}, \bibinfo{person}{Hongxia Jin}, {and} \bibinfo{person}{Zsolt Kira}.}
  \bibinfo{year}{2020}\natexlab{}.
\newblock \showarticletitle{Generalized odin: Detecting out-of-distribution
  image without learning from out-of-distribution data}. In
  \bibinfo{booktitle}{\emph{Proceedings of the IEEE/CVF Conference on Computer
  Vision and Pattern Recognition}}. \bibinfo{pages}{10951--10960}.
\newblock


\bibitem[\protect\citeauthoryear{Ji, Luo, Sun, Chen, Luo, Wu, Gao, and Ji}{Ji
  et~al\mbox{.}}{2021}]%
        {ji2021improving}
\bibfield{author}{\bibinfo{person}{Jiayi Ji}, \bibinfo{person}{Yunpeng Luo},
  \bibinfo{person}{Xiaoshuai Sun}, \bibinfo{person}{Fuhai Chen},
  \bibinfo{person}{Gen Luo}, \bibinfo{person}{Yongjian Wu},
  \bibinfo{person}{Yue Gao}, {and} \bibinfo{person}{Rongrong Ji}.}
  \bibinfo{year}{2021}\natexlab{}.
\newblock \showarticletitle{Improving image captioning by leveraging intra-and
  inter-layer global representation in transformer network}. In
  \bibinfo{booktitle}{\emph{Proceedings of the AAAI Conference on Artificial
  Intelligence}}, Vol.~\bibinfo{volume}{35}. \bibinfo{pages}{1655--1663}.
\newblock


\bibitem[\protect\citeauthoryear{Jung, Wada, Crall, Tanaka, Graving,
  et~al\mbox{.}}{Jung et~al\mbox{.}}{2019}]%
        {imgaug}
\bibfield{author}{\bibinfo{person}{Alexander~B. Jung}, \bibinfo{person}{Kentaro
  Wada}, \bibinfo{person}{Jon Crall}, \bibinfo{person}{Satoshi Tanaka},
  \bibinfo{person}{Jake Graving}, {et~al\mbox{.}}}
  \bibinfo{year}{2019}\natexlab{}.
\newblock \bibinfo{title}{{imgaug}}.
\newblock \bibinfo{howpublished}{https://github.com/aleju/imgaug}.
\newblock
\newblock
\shownote{Online; accessed 25-Sept-2019.}


\bibitem[\protect\citeauthoryear{Koner, Sinhamahapatra, Roscher, G{\"u}nnemann,
  and Tresp}{Koner et~al\mbox{.}}{2021}]%
        {koner2021oodformer}
\bibfield{author}{\bibinfo{person}{Rajat Koner}, \bibinfo{person}{Poulami
  Sinhamahapatra}, \bibinfo{person}{Karsten Roscher}, \bibinfo{person}{Stephan
  G{\"u}nnemann}, {and} \bibinfo{person}{Volker Tresp}.}
  \bibinfo{year}{2021}\natexlab{}.
\newblock \showarticletitle{Oodformer: Out-of-distribution detection
  transformer}.
\newblock \bibinfo{journal}{\emph{BMVC}} (\bibinfo{year}{2021}).
\newblock


\bibitem[\protect\citeauthoryear{Krasin, Duerig, Alldrin, Ferrari,
  Abu-El-Haija, Kuznetsova, Rom, Uijlings, Popov, Veit, et~al\mbox{.}}{Krasin
  et~al\mbox{.}}{2017}]%
        {krasin2017openimages}
\bibfield{author}{\bibinfo{person}{Ivan Krasin}, \bibinfo{person}{Tom Duerig},
  \bibinfo{person}{Neil Alldrin}, \bibinfo{person}{Vittorio Ferrari},
  \bibinfo{person}{Sami Abu-El-Haija}, \bibinfo{person}{Alina Kuznetsova},
  \bibinfo{person}{Hassan Rom}, \bibinfo{person}{Jasper Uijlings},
  \bibinfo{person}{Stefan Popov}, \bibinfo{person}{Andreas Veit},
  {et~al\mbox{.}}} \bibinfo{year}{2017}\natexlab{}.
\newblock \showarticletitle{Openimages: A public dataset for large-scale
  multi-label and multi-class image classification}.
\newblock \bibinfo{journal}{\emph{Dataset available from https://github.
  com/openimages}} \bibinfo{volume}{2}, \bibinfo{number}{3}
  (\bibinfo{year}{2017}), \bibinfo{pages}{2--3}.
\newblock


\bibitem[\protect\citeauthoryear{Kuznetsova, Ordonez, Berg, Berg, and
  Choi}{Kuznetsova et~al\mbox{.}}{2012}]%
        {kuznetsova2012collective}
\bibfield{author}{\bibinfo{person}{Polina Kuznetsova}, \bibinfo{person}{Vicente
  Ordonez}, \bibinfo{person}{Alexander~C Berg}, \bibinfo{person}{Tamara~L
  Berg}, {and} \bibinfo{person}{Yejin Choi}.} \bibinfo{year}{2012}\natexlab{}.
\newblock \showarticletitle{Collective generation of natural image
  descriptions}. In \bibinfo{booktitle}{\emph{Proceedings of the 50th Annual
  Meeting of the Association for Computational Linguistics: Long Papers-Volume
  1}}. Association for Computational Linguistics, \bibinfo{pages}{359--368}.
\newblock


\bibitem[\protect\citeauthoryear{Lee, Lee, Lee, and Shin}{Lee
  et~al\mbox{.}}{2018a}]%
        {lee2017training}
\bibfield{author}{\bibinfo{person}{Kimin Lee}, \bibinfo{person}{Honglak Lee},
  \bibinfo{person}{Kibok Lee}, {and} \bibinfo{person}{Jinwoo Shin}.}
  \bibinfo{year}{2018}\natexlab{a}.
\newblock \showarticletitle{Training confidence-calibrated classifiers for
  detecting out-of-distribution samples}.
\newblock \bibinfo{journal}{\emph{Proceedings of International Conference on
  Learning Representations}} (\bibinfo{year}{2018}).
\newblock


\bibitem[\protect\citeauthoryear{Lee, Lee, Lee, and Shin}{Lee
  et~al\mbox{.}}{2018b}]%
        {lee2018simple}
\bibfield{author}{\bibinfo{person}{Kimin Lee}, \bibinfo{person}{Kibok Lee},
  \bibinfo{person}{Honglak Lee}, {and} \bibinfo{person}{Jinwoo Shin}.}
  \bibinfo{year}{2018}\natexlab{b}.
\newblock \showarticletitle{A simple unified framework for detecting
  out-of-distribution samples and adversarial attacks}.
\newblock \bibinfo{journal}{\emph{Advances in neural information processing
  systems}}  \bibinfo{volume}{31} (\bibinfo{year}{2018}).
\newblock


\bibitem[\protect\citeauthoryear{Li, Zhu, Liu, and Yang}{Li
  et~al\mbox{.}}{2019}]%
        {li2019entangled}
\bibfield{author}{\bibinfo{person}{Guang Li}, \bibinfo{person}{Linchao Zhu},
  \bibinfo{person}{Ping Liu}, {and} \bibinfo{person}{Yi Yang}.}
  \bibinfo{year}{2019}\natexlab{}.
\newblock \showarticletitle{Entangled transformer for image captioning}. In
  \bibinfo{booktitle}{\emph{Proceedings of the IEEE/CVF International
  Conference on Computer Vision}}. \bibinfo{pages}{8928--8937}.
\newblock


\bibitem[\protect\citeauthoryear{Li, Li, Sun, Fan, Zhang, Wu, Meng, and
  Zhang}{Li et~al\mbox{.}}{2021}]%
        {li2021k}
\bibfield{author}{\bibinfo{person}{Xiaoya Li}, \bibinfo{person}{Jiwei Li},
  \bibinfo{person}{Xiaofei Sun}, \bibinfo{person}{Chun Fan},
  \bibinfo{person}{Tianwei Zhang}, \bibinfo{person}{Fei Wu},
  \bibinfo{person}{Yuxian Meng}, {and} \bibinfo{person}{Jun Zhang}.}
  \bibinfo{year}{2021}\natexlab{}.
\newblock \showarticletitle{$ k $ Folden: $ k $-Fold Ensemble for
  Out-Of-Distribution Detection}.
\newblock \bibinfo{journal}{\emph{EMNLP}} (\bibinfo{year}{2021}).
\newblock


\bibitem[\protect\citeauthoryear{Liang, Li, and Srikant}{Liang
  et~al\mbox{.}}{2018}]%
        {liang2017enhancing}
\bibfield{author}{\bibinfo{person}{Shiyu Liang}, \bibinfo{person}{Yixuan Li},
  {and} \bibinfo{person}{Rayadurgam Srikant}.} \bibinfo{year}{2018}\natexlab{}.
\newblock \showarticletitle{Enhancing the reliability of out-of-distribution
  image detection in neural networks}.
\newblock \bibinfo{journal}{\emph{Proceedings of International Conference on
  Learning Representations}} (\bibinfo{year}{2018}).
\newblock


\bibitem[\protect\citeauthoryear{Lin, Maire, Belongie, Hays, Perona, Ramanan,
  Doll{\'a}r, and Zitnick}{Lin et~al\mbox{.}}{2014}]%
        {lin2014microsoft}
\bibfield{author}{\bibinfo{person}{Tsung-Yi Lin}, \bibinfo{person}{Michael
  Maire}, \bibinfo{person}{Serge Belongie}, \bibinfo{person}{James Hays},
  \bibinfo{person}{Pietro Perona}, \bibinfo{person}{Deva Ramanan},
  \bibinfo{person}{Piotr Doll{\'a}r}, {and} \bibinfo{person}{C~Lawrence
  Zitnick}.} \bibinfo{year}{2014}\natexlab{}.
\newblock \showarticletitle{Microsoft coco: Common objects in context}. In
  \bibinfo{booktitle}{\emph{European conference on computer vision}}. Springer,
  \bibinfo{pages}{740--755}.
\newblock


\bibitem[\protect\citeauthoryear{Lu, Yang, Batra, and Parikh}{Lu
  et~al\mbox{.}}{2018}]%
        {lu2018neural}
\bibfield{author}{\bibinfo{person}{Jiasen Lu}, \bibinfo{person}{Jianwei Yang},
  \bibinfo{person}{Dhruv Batra}, {and} \bibinfo{person}{Devi Parikh}.}
  \bibinfo{year}{2018}\natexlab{}.
\newblock \showarticletitle{Neural baby talk}. In
  \bibinfo{booktitle}{\emph{Proceedings of the IEEE Conference on Computer
  Vision and Pattern Recognition}}. \bibinfo{pages}{7219--7228}.
\newblock


\bibitem[\protect\citeauthoryear{MacLeod, Bennett, Morris, and Cutrell}{MacLeod
  et~al\mbox{.}}{2017}]%
        {macleod2017understanding}
\bibfield{author}{\bibinfo{person}{Haley MacLeod}, \bibinfo{person}{Cynthia~L
  Bennett}, \bibinfo{person}{Meredith~Ringel Morris}, {and}
  \bibinfo{person}{Edward Cutrell}.} \bibinfo{year}{2017}\natexlab{}.
\newblock \showarticletitle{Understanding blind people's experiences with
  computer-generated captions of social media images}. In
  \bibinfo{booktitle}{\emph{Proceedings of the 2017 CHI Conference on Human
  Factors in Computing Systems}}. ACM, \bibinfo{pages}{5988--5999}.
\newblock


\bibitem[\protect\citeauthoryear{Nguyen, Yosinski, and Clune}{Nguyen
  et~al\mbox{.}}{2015}]%
        {nguyen2015deep}
\bibfield{author}{\bibinfo{person}{Anh Nguyen}, \bibinfo{person}{Jason
  Yosinski}, {and} \bibinfo{person}{Jeff Clune}.}
  \bibinfo{year}{2015}\natexlab{}.
\newblock \showarticletitle{Deep neural networks are easily fooled: High
  confidence predictions for unrecognizable images}. In
  \bibinfo{booktitle}{\emph{Proceedings of the IEEE conference on computer
  vision and pattern recognition}}. \bibinfo{pages}{427--436}.
\newblock


\bibitem[\protect\citeauthoryear{Papineni, Roukos, Ward, and Zhu}{Papineni
  et~al\mbox{.}}{2002}]%
        {papineni2002bleu}
\bibfield{author}{\bibinfo{person}{Kishore Papineni}, \bibinfo{person}{Salim
  Roukos}, \bibinfo{person}{Todd Ward}, {and} \bibinfo{person}{Wei-Jing Zhu}.}
  \bibinfo{year}{2002}\natexlab{}.
\newblock \showarticletitle{BLEU: a method for automatic evaluation of machine
  translation}. In \bibinfo{booktitle}{\emph{Proceedings of the 40th annual
  meeting on association for computational linguistics}}. Association for
  Computational Linguistics, \bibinfo{pages}{311--318}.
\newblock


\bibitem[\protect\citeauthoryear{Reyes-Aldasoro and Bhalerao}{Reyes-Aldasoro
  and Bhalerao}{2006}]%
        {reyes2006bhattacharyya}
\bibfield{author}{\bibinfo{person}{Constantino~Carlos Reyes-Aldasoro} {and}
  \bibinfo{person}{Abhir Bhalerao}.} \bibinfo{year}{2006}\natexlab{}.
\newblock \showarticletitle{The Bhattacharyya space for feature selection and
  its application to texture segmentation}.
\newblock \bibinfo{journal}{\emph{Pattern Recognition}} \bibinfo{volume}{39},
  \bibinfo{number}{5} (\bibinfo{year}{2006}), \bibinfo{pages}{812--826}.
\newblock


\bibitem[\protect\citeauthoryear{Sadeh, Fritz, Shalev, and Oks}{Sadeh
  et~al\mbox{.}}{2019a}]%
        {sadeh2019generating}
\bibfield{author}{\bibinfo{person}{Gil Sadeh}, \bibinfo{person}{Lior Fritz},
  \bibinfo{person}{Gabi Shalev}, {and} \bibinfo{person}{Eduard Oks}.}
  \bibinfo{year}{2019}\natexlab{a}.
\newblock \showarticletitle{Generating Diverse and Informative Natural Language
  Fashion Feedback}.
\newblock \bibinfo{journal}{\emph{CVPR Workshop on Language and Vision}}
  (\bibinfo{year}{2019}).
\newblock


\bibitem[\protect\citeauthoryear{Sadeh, Fritz, Shalev, and Oks}{Sadeh
  et~al\mbox{.}}{2019b}]%
        {sadeh2019joint}
\bibfield{author}{\bibinfo{person}{Gil Sadeh}, \bibinfo{person}{Lior Fritz},
  \bibinfo{person}{Gabi Shalev}, {and} \bibinfo{person}{Eduard Oks}.}
  \bibinfo{year}{2019}\natexlab{b}.
\newblock \showarticletitle{Joint visual-textual embedding for multimodal style
  search}.
\newblock \bibinfo{journal}{\emph{CVPR Workshop on Language and Vision}}
  (\bibinfo{year}{2019}).
\newblock


\bibitem[\protect\citeauthoryear{Shahid, Rappon, and Berta}{Shahid
  et~al\mbox{.}}{2019}]%
        {shahid2019applications}
\bibfield{author}{\bibinfo{person}{Nida Shahid}, \bibinfo{person}{Tim Rappon},
  {and} \bibinfo{person}{Whitney Berta}.} \bibinfo{year}{2019}\natexlab{}.
\newblock \showarticletitle{Applications of artificial neural networks in
  health care organizational decision-making: A scoping review}.
\newblock \bibinfo{journal}{\emph{PloS one}} \bibinfo{volume}{14},
  \bibinfo{number}{2} (\bibinfo{year}{2019}).
\newblock


\bibitem[\protect\citeauthoryear{Shalev, Adi, and Keshet}{Shalev
  et~al\mbox{.}}{2018}]%
        {shalev2018out}
\bibfield{author}{\bibinfo{person}{Gabi Shalev}, \bibinfo{person}{Yossi Adi},
  {and} \bibinfo{person}{Joseph Keshet}.} \bibinfo{year}{2018}\natexlab{}.
\newblock \showarticletitle{Out-of-distribution detection using multiple
  semantic label representations}. In \bibinfo{booktitle}{\emph{Advances in
  Neural Information Processing Systems}}. \bibinfo{pages}{7375--7385}.
\newblock


\bibitem[\protect\citeauthoryear{Shalev, Shalev, and Keshet}{Shalev
  et~al\mbox{.}}{2020}]%
        {shalev2020redesigning}
\bibfield{author}{\bibinfo{person}{Gabi Shalev}, \bibinfo{person}{Gal-Lev
  Shalev}, {and} \bibinfo{person}{Joseph Keshet}.}
  \bibinfo{year}{2020}\natexlab{}.
\newblock \showarticletitle{Redesigning the classification layer by randomizing
  the class representation vectors}.
\newblock \bibinfo{journal}{\emph{arXiv preprint arXiv:2011.08704}}
  (\bibinfo{year}{2020}).
\newblock


\bibitem[\protect\citeauthoryear{Shalev, Shalev, and Keshet}{Shalev
  et~al\mbox{.}}{2021}]%
        {shalev2021randomized}
\bibfield{author}{\bibinfo{person}{Gal-Lev Shalev}, \bibinfo{person}{Gabi
  Shalev}, {and} \bibinfo{person}{Joseph Keshet}.}
  \bibinfo{year}{2021}\natexlab{}.
\newblock \showarticletitle{On Randomized Classification Layers and Their
  Implications in Natural Language Generation}. In
  \bibinfo{booktitle}{\emph{Proceedings of the Third Workshop on Multimodal
  Artificial Intelligence}}. \bibinfo{pages}{6--11}.
\newblock


\bibitem[\protect\citeauthoryear{Tran, He, Zhang, Sun, Carapcea, Thrasher,
  Buehler, and Sienkiewicz}{Tran et~al\mbox{.}}{2016}]%
        {tran2016rich}
\bibfield{author}{\bibinfo{person}{Kenneth Tran}, \bibinfo{person}{Xiaodong
  He}, \bibinfo{person}{Lei Zhang}, \bibinfo{person}{Jian Sun},
  \bibinfo{person}{Cornelia Carapcea}, \bibinfo{person}{Chris Thrasher},
  \bibinfo{person}{Chris Buehler}, {and} \bibinfo{person}{Chris Sienkiewicz}.}
  \bibinfo{year}{2016}\natexlab{}.
\newblock \showarticletitle{Rich image captioning in the wild}. In
  \bibinfo{booktitle}{\emph{Proceedings of the IEEE Conference on Computer
  Vision and Pattern Recognition Workshops}}. \bibinfo{pages}{49--56}.
\newblock


\bibitem[\protect\citeauthoryear{Vedantam, Lawrence~Zitnick, and
  Parikh}{Vedantam et~al\mbox{.}}{2015}]%
        {vedantam2015cider}
\bibfield{author}{\bibinfo{person}{Ramakrishna Vedantam}, \bibinfo{person}{C
  Lawrence~Zitnick}, {and} \bibinfo{person}{Devi Parikh}.}
  \bibinfo{year}{2015}\natexlab{}.
\newblock \showarticletitle{Cider: Consensus-based image description
  evaluation}. In \bibinfo{booktitle}{\emph{Proceedings of the IEEE conference
  on computer vision and pattern recognition}}. \bibinfo{pages}{4566--4575}.
\newblock


\bibitem[\protect\citeauthoryear{Venkatakrishnan, Kim, Eisawy, Pfister, and
  Navab}{Venkatakrishnan et~al\mbox{.}}{2020}]%
        {venkatakrishnan2020self}
\bibfield{author}{\bibinfo{person}{Abinav~Ravi Venkatakrishnan},
  \bibinfo{person}{Seong~Tae Kim}, \bibinfo{person}{Rami Eisawy},
  \bibinfo{person}{Franz Pfister}, {and} \bibinfo{person}{Nassir Navab}.}
  \bibinfo{year}{2020}\natexlab{}.
\newblock \showarticletitle{Self-supervised out-of-distribution detection in
  brain CT scans}.
\newblock \bibinfo{journal}{\emph{arXiv preprint arXiv:2011.05428}}
  (\bibinfo{year}{2020}).
\newblock


\bibitem[\protect\citeauthoryear{Vinyals, Toshev, Bengio, and Erhan}{Vinyals
  et~al\mbox{.}}{2015}]%
        {vinyals2015show}
\bibfield{author}{\bibinfo{person}{Oriol Vinyals}, \bibinfo{person}{Alexander
  Toshev}, \bibinfo{person}{Samy Bengio}, {and} \bibinfo{person}{Dumitru
  Erhan}.} \bibinfo{year}{2015}\natexlab{}.
\newblock \showarticletitle{Show and tell: A neural image caption generator}.
  In \bibinfo{booktitle}{\emph{Proceedings of the IEEE conference on computer
  vision and pattern recognition}}. \bibinfo{pages}{3156--3164}.
\newblock


\end{thebibliography}
\appendix
\subsection{Unknown Objects}
For creating the \textit{unknown object} and the \textit{cropped
unknown object} sets we collected images containing 38 objects which are not appearing in COCO
training sets. The objects we fetched are: \textit{bomb,
camel, calculator, dolphin, lion, beetle, chime,
dumbbell, hammer, belt, alpaca, dice, balloon,
frog, cheetah, hippopotamus, fork, fax, kangaroo,
flute, cart, binoculars, insect, armadillo, helicopter,
grape, coin, handgun, otter, beaker, bat, bee, lizard,
hamster, cello, axe, barrel, egg}. Examples can be
seen in Fig-\ref{fig:samples}. Images are sourced from
open images V4.

\subsection{Baseline Model}
The baseline model was trained on COCO dataset labels: \textit{person, umbrella, tie, backpack, handbag, suitcase, bicycle, motorcycle, bus, truck, truck, airplane, train,
boat, bench, stop sign, traffic light, fire hydrant,
parking meter, zebra, elephant, sheep, dog, bird,
cat, horse, cow, bear, giraffe, surfboard, baseball
glove, kite, snowboard, frisbee, skis, sports ball,
baseball bat, skateboard, tennis racket, bowl, knife,
cup, bottle, wine glass, fork, spoon, donut, hot dog,
broccoli, sandwich, banana, apple, orange, carrot,
pizza, cake, dining table, potted plant, chair, couch,
bed, toilet, keyboard, mouse, tv, laptop, remote, cell phone, refrigerator, toaster, microwave, oven, sink, toothbrush, teddy bear, vase, book, clock, scissors,} and \textit{hair drier}.
\end{document}